\RequirePackage[OT1]{fontenc}
\documentclass[letterpaper, 10 pt, conference]{activewam}  

\IEEEoverridecommandlockouts                              
\usepackage{graphics}
\usepackage{graphicx}
\usepackage[tight,footnotesize]{subfigure}
\usepackage{multirow}
\usepackage[cmex10]{amsmath}
\usepackage{amssymb}
\usepackage{amsfonts}
\usepackage{booktabs}
\usepackage{arydshln}
\usepackage{textcomp}
\usepackage{url}
\usepackage{color}
\usepackage[table]{xcolor}

\usepackage{soul}
\usepackage{xcolor}
\usepackage{colortbl}
\usepackage{wrapfig}
\usepackage[english]{babel}
\usepackage[small]{caption}
\usepackage{capt-of}

\providecommand{\awSD}[2]{}

\renewcommand{\awSD}[2]{\mbox{#1{\fontsize{6}{7}\selectfont\,\ensuremath{\pm#2}}}}



\usepackage[hidelinks,bookmarks=false]{hyperref}
\usepackage{balance}
\ifdefined\pdfminorversion
\else
  \AtBeginDvi{}
\fi
\usepackage[protrusion=false]{microtype}
\usepackage{placeins}

\hypersetup{pdftitle={ActiveWAM},pdfauthor={Renjun Wu, Luzhou Ge, Xuesong Li},pdfsubject={Active-view world-action modeling for robot manipulation}}
\newcommand{\methodname}{ActiveWAM}
\title{\LARGE \bf ActiveWAM: Evidence-Aware Active Vision for World-Action Models}
\usepackage{marvosym}
\author{\textbf{Renjun Wu}\textsuperscript{1*},
\textbf{Luzhou Ge}\textsuperscript{1*},
\textbf{Xuesong Li}\textsuperscript{1\Letter}\\
\textsuperscript{1}Beijing Institute of Technology\\
\normalsize\textsuperscript{*}Equal contribution, \textsuperscript{\Letter}Corresponding author\\
\normalsize\textcolor[rgb]{0,0,0.45}{\url{https://icr-lab.github.io/ActiveWAM}}}

\makeatletter
\let\@oldmaketitle\@maketitle
\renewcommand{\@maketitle}{\@oldmaketitle
  \begin{center}
    \includegraphics[width=\textwidth]{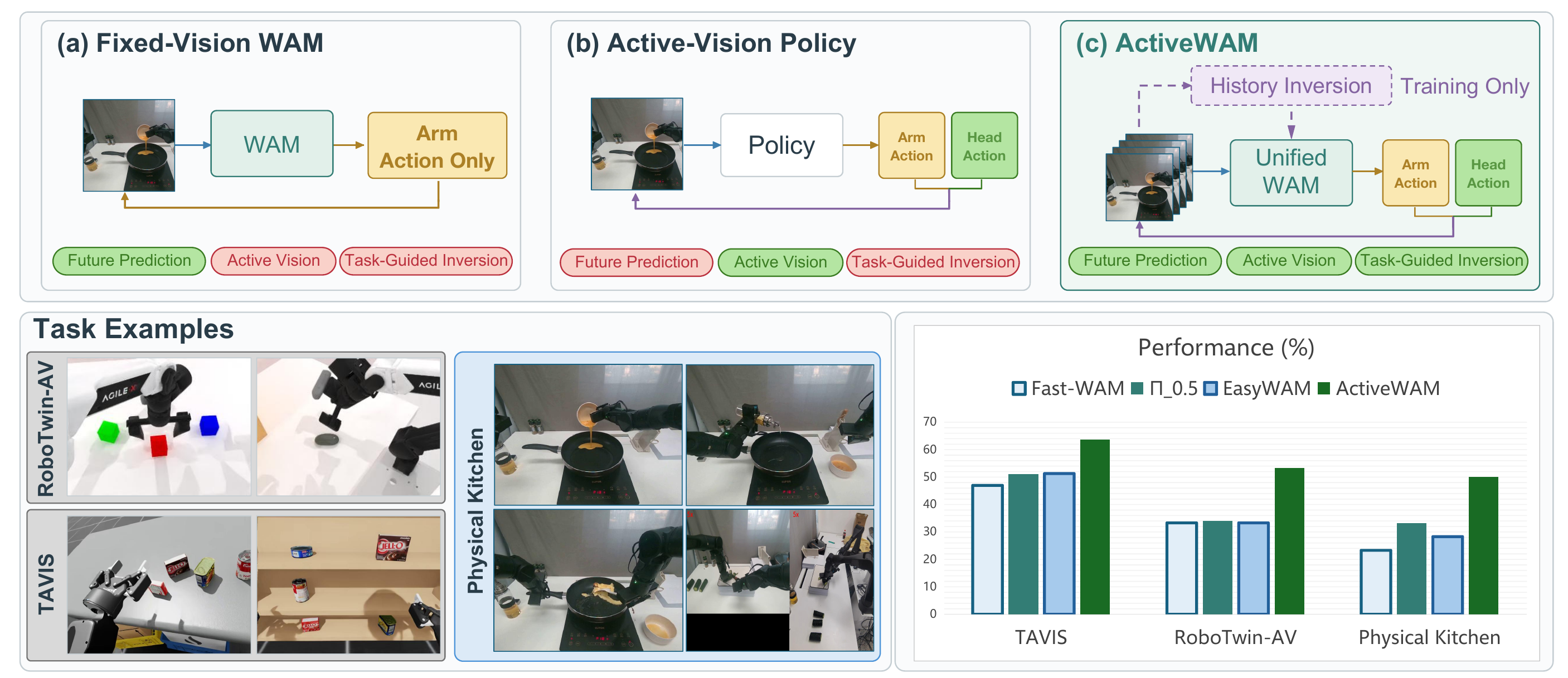}
    \captionof{figure}{\textbf{(a)} Fixed-vision WAM generates arm actions from static camera observations.
\textbf{(b)} Active-vision policy controls head separately from manipulation.
\textbf{(c)} ActiveWAM unifies head and arm generation in a common world-action model, using training-time history inversion to preserve task-relevant evidence while learning executable pan/tilt control.
Task examples: simulation (RoboTwin-AV) shows object placement requiring spatial reframing; physical kitchen setup demonstrates composed multi-stage manipulation.
Bar charts: ActiveWAM improves TAVIS success by 17.0pp on Head/GR1T2, RoboTwin-AV compound success by 20.0pp over Fast-WAM, and physical success by 26.7pp over Fast-WAM.}\label{fig:teaser}\end{center}
  \setcounter{figure}{1}\vspace{-1em}
}
\makeatother

\begin{document}
\raggedbottom

\maketitle
\thispagestyle{empty}
\pagestyle{empty}

\bstctlcite{IEEEexample:BSTcontrol}

\begin{abstract}
Active vision manipulation requires a policy to control both its camera and its
end-effectors, yet camera motion determines which evidence remains visible within
finite observation windows. Acquiring a new view can displace task-critical cues,
while retaining a view forgoes potentially useful observations. We formulate this
as an evidence-aware retain--acquire problem and present ActiveWAM, a unified
world--action model that learns observation and manipulation jointly. To this end,
we propose training-time inversion which constrains a frozen video prior by
task-bearing source evidence and visible temporal changes, eliminating the need
for test-time inversion or candidate ranking. At deployment, the policy generates
bimanual and pan/tilt actions—including stay and reacquisition behaviors—from
view-aware history, and updates context from newly measured RGB observations.
Future-video prediction serves as a co-training signal, while action generation
requires neither future-video decoding nor optimal viewpoint annotations. We
introduce RoboTwin-AV, a 50-task benchmark with executable pan/tilt control and
automatically generated demonstrations. ActiveWAM improves TAVIS out-of-distribution
success by up to 17.0 percentage points over the strongest baselines, achieves 20.0
additional points over Fast-WAM on RoboTwin-AV, and outperforms it by 26.7 points on real-world physical kitchen tasks.
\end{abstract}

\section{Introduction}
\label{sec:introduction}

General-purpose manipulation requires robots to both acquire task-relevant visual evidence and act on it. Vision--Language--Action (VLA) policies connect observations and instructions to commands~\cite{openvla2024,pi02024}, while World--Action Models (WAMs) additionally learn the visual consequences of those commands~\cite{yuan2026fastwam,cosmospolicy2026,easywam2026}. In bimanual manipulation, coordinated arm actions change the scene while camera actions determine which spatial relations remain visible within the policy's finite observation window.

Useful observation is not equivalent to removing occlusion. During approach, a view should expose object--end-effector relations; during placement, the destination becomes equally important. Even in an unobstructed workspace, moving a well-positioned camera can discard context. The appropriate observation behavior includes retaining a good view, not only searching for a better one. We call this the \emph{retain--acquire} problem: a policy must preserve task-critical evidence in its recent representation while acquiring or reacquiring physical evidence when the current view is insufficient.

Recent benchmarks demonstrate that active camera control depends on task structure rather than universal heuristics. AV-ALOHA evaluates active, static, and wrist cameras in bimanual tasks~\cite{avaloha2024}; TAVIS studies conditional selection, search, and waiting behaviors under distribution shifts~\cite{tavis2026}; AVR combines viewpoint and focal-length control~\cite{avr2025}; See2Act couples action refinement to target-centric views~\cite{see2act2026}. These results establish that observation control cannot rely on fixed rules and expose a temporal challenge: changing the view can acquire new evidence while removing previously useful evidence.

Pretrained representations~\cite{r3m2022,mvp2023} and domain randomization~\cite{dr2017} support visual transfer, yet history still mixes task evidence with appearance and camera motion. Stem-OB uses diffusion inversion to improve visual generalization~\cite{stemob2024}, but a generic transform neither specifies which historical evidence must survive nor preserves temporal changes that can overturn earlier decisions.

We address the joint question: \emph{how can a world--action model learn to retain task-relevant evidence while controlling which evidence it acquires next?} Our key insight is to couple a task-selective training transformation with executable observation actions through the same demonstrated continuation. The transformation changes the representation of an observed history; head and arm actions change the physical observation stream. Neither substitutes for the other: transforming a history cannot reveal an unobserved surface, and moving a camera cannot by itself ensure that a previously seen cue remains available.

We present \methodname{}, an evidence-aware active vision WAM that combines task-guided history inversion with unified head--arm generation (Fig.~\ref{fig:teaser}). Training-Time Inversion (TTI) uses a frozen video prior constrained by task-bearing source evidence and visible temporal changes. A view-aware history adapter represents camera identity, pose, exposure time, and validity alongside current RGB. The policy learns from raw and transformed conditions with identical original targets. Future-video prediction provides a co-training signal; deployment uses raw observations to generate one head--arm trajectory, without online inversion, candidate ranking, or manually annotated optimal viewpoints.

To evaluate this retain--acquire coupling at bimanual-task scale, we introduce \emph{RoboTwin-AV}, an active vision benchmark extension of RoboTwin 2.0~\cite{robotwin22025} retaining all 50 task definitions while adding executable pan/tilt control, synchronized RGB, and automatically generated demonstrations. We evaluate on the independently released TAVIS benchmark and use RoboTwin-AV for large-scale controlled comparisons, complemented by fixed-camera generalization tests and physical trials.

Our contributions are threefold:
\begin{itemize}
\item We formulate active vision manipulation as evidence-aware retain--acquire behavior and introduce task-guided inversion that preserves task evidence and visible temporal changes within a finite history window while transferring the training signal to a raw-input policy.
\item We propose ActiveWAM, a unified policy that generates executable head--arm actions from view-aware history, including stay and reacquisition behaviors; future-video prediction is used as a co-training signal rather than a required test-time output.
\item We introduce RoboTwin-AV, a 50-task benchmark extension with executable pan/tilt control, and evaluate the retain--acquire interaction through factorial control, collector-shift tests, external baselines, and physical bimanual manipulation.
\end{itemize}

\begin{figure*}[t]
 \centering
 \includegraphics[width=\textwidth]{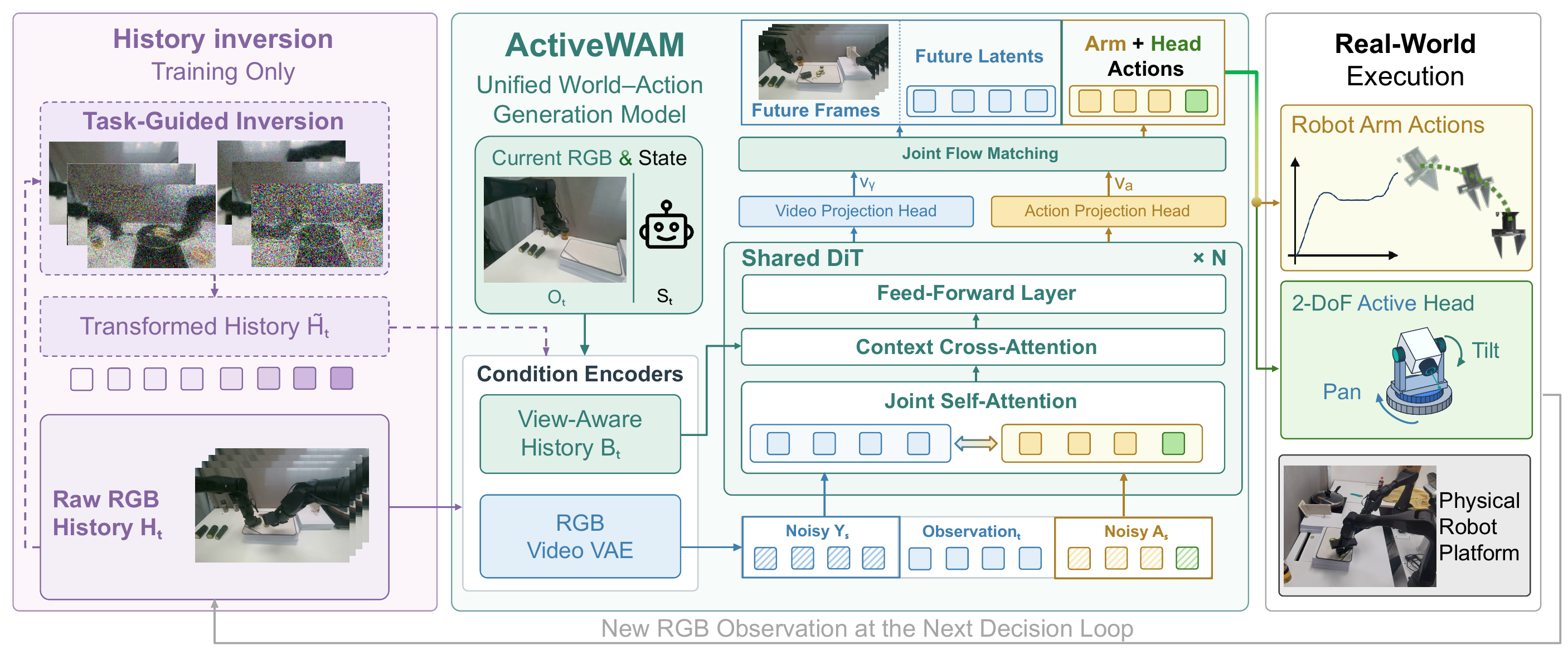}
 \caption{ActiveWAM overview. Active vision manipulation is treated as evidence-aware retain--acquire control. Training-time source-constrained inversion transforms observed histories, while shared WAM passes receive raw or transformed conditions with identical original future/action targets. View-aware history and current RGB condition encoders supply recent evidence and its camera context; the action branch determines executable arm and head control, with future-video prediction used as a co-training signal. Deployment retains raw inputs, one action-generation pass, and actual new measurements after a short executed prefix. It uses neither inversion nor candidate ranking, and requires no manually annotated optimal viewpoints or information-gain objective.}
 \label{fig:overview}
\end{figure*}

\section{Related Work}
\label{sec:related}

\subsection{Active Observation and Camera Control}

Active cameras can expose occluded cues and track moving objects, yet each view change replaces previously visible information with new observations. This creates a retain--acquire trade-off: acquiring new evidence competes with retaining prior cues within finite history windows. AV-ALOHA evaluates active, static, and wrist cameras across bimanual tasks~\cite{avaloha2024}. TAVIS introduces task-dependent gaze, waiting, search, and exploration under distribution shifts~\cite{tavis2026}. AVR combines viewpoint and focal-length control~\cite{avr2025}. These results establish that observation control must adapt to the evidence each manipulation stage requires.

Existing learned approaches treat observation and manipulation separately. Sun et al.\ decouple gaze via look-at inverse kinematics, reducing policy complexity but preventing joint reasoning over camera and arm consequences~\cite{lookat2024}. See2Act couples action refinement with test-time viewpoint search, incurring computational overhead and lacking the closed-loop reactivity manipulation requires~\cite{see2act2026}. Decoupled control cannot capture joint consequences of camera and arm actions. We address this through evidence-aware retain--acquire generation within one world--action model trained end-to-end, using training-time history inversion to transfer evidence retention to raw-input deployment.

\subsection{World--Action Models and Unified Generation}
Generalist policies scale robot learning through shared transformers~\cite{rt12022,octo2024,pi052025}. World--Action Models (WAMs) extend this paradigm by jointly generating future-video predictions and executable actions. DreamZero and DiT4DiT couple video generation with action outputs~\cite{ye2026worldactionmodelszeroshot,ma2026dit4dit}. EasyWAM-Unified provides a unified transformer interface for joint future/action generation~\cite{easywam2026}, while Fast-WAM demonstrates that video prediction can serve as a co-training signal without requiring future-video decoding at test time~\cite{yuan2026fastwam}. We build on this unified-generation framework, extending it to couple task-selective history transformation with physical sensing actions. Following Fast-WAM, we evaluate video co-training and action-only inference roles separately.

\subsection{Observation Inversion and Temporal Consistency}
Diffusion-based inversion enables deterministic reconstruction along learned trajectories~\cite{ddim2020,nulltext2023}. Video editing methods extend this principle through source-guided transformations, using attention control and inter-frame feature correspondences to maintain temporal consistency~\cite{videop2p2023,fatezero2023,tokenflow2023}. In robotics, Stem-OB applies observation inversion to transfer transformed-input training benefits to raw-image deployment~\cite{stemob2024}. However, Stem-OB operates on single current observations and does not constrain which historical evidence must survive view transformations. We extend this principle to observed video history, constraining partial history transformation by task-bearing evidence and visible temporal changes while retaining the original action continuation and future outcome as targets.

\setcounter{section}{2}
\section{Method}
\label{sec:method}
\subsection{Problem Formulation and Overview}
\label{sec:formulation}

We denote the observation history from camera $c$ at decision time $t$ as 
$\mathcal H_t^c=(I^c_{t_1},\ldots,I^c_{t_L})$, containing only frames already 
exposed by time $t$. The policy condition 
$C_t=(\ell,\mathcal H_t,s_t,p_{\leq t},\kappa,\delta_{\leq t})$ comprises the 
language instruction, multi-camera RGB history, robot state, measured camera poses, 
calibration, and frame acquisition metadata.The policy uses a recent window of observed frames, together with acquisition metadata that records their age and validity.

Our physical setup consists of a fixed-base dual-arm robot with a single RGB camera 
on a two-DoF pan/tilt head at fixed focal length. The action space comprises six 
pose increments and one gripper command per arm, plus two head increments $\Delta p_t$:
\begin{equation}
 a_t=[\Delta x_t^l,\Delta\omega_t^l,g_t^l,
      \Delta x_t^r,\Delta\omega_t^r,g_t^r,\Delta p_t]\in\mathbb R^{16}.
 \label{eq:action}
\end{equation}
Zero head motion is a valid \emph{stay} action, allowing the policy to retain a 
stable view when the current observation already exposes task-relevant cues. We 
model a single conditional distribution
\begin{equation}
 p_\theta(Y_{t+1:t+K},A_{t:t+H_a-1}\mid C_t)
 \label{eq:problem}
\end{equation}
over view-specific future-video latents $Y$ and executable action chunks $A$. 
This unified formulation couples future-video prediction and action generation: 
head rotation changes both the field of view and task framing.

\paragraph*{Overview}
Fig.~\ref{fig:overview} illustrates ActiveWAM's three core mechanisms. 
Training-time task-guided history inversion (Sec.~\ref{sec:inversion}) preserves 
task-bearing evidence and visible temporal changes. A view-aware history adapter 
(Sec.~\ref{sec:joint_model}) conditions a shared world--action generator on 
available evidence. Paired learning (Sec.~\ref{sec:execution}) anchors raw and 
transformed conditions to identical original targets. Deployment uses raw history, 
generates one joint action trajectory, and updates context from actual new 
measurements after executing a short prefix.

\subsection{Task-Guided History Inversion}
\label{sec:inversion}

Active vision manipulation introduces a dual challenge: head actions change both
\emph{what} is observed and \emph{where} task-critical cues appear. To address
retention, we train the policy on transformed histories that preserve
task-relevant evidence, teaching it which cues must survive view changes.

We extend observation inversion~\cite{stemob2024} to \emph{observed video history}.
A frozen Wan T2V video prior~\cite{wan2025} processes each camera's history
independently. Cameras are not temporal neighbors, and the prior receives neither
future observations nor task-privileged information. Bidirectional processing within
$\mathcal H_t$ remains causal for decision $t$ because every frame has already been
observed. The prior supplies a source-constrained transformation direction, while
the recorded action continuation and future outcome remain fixed, ensuring that
inversion introduces no new physical evidence. Training-Time Inversion (TTI)
generates paired examples for learning, while deployment uses raw history.

\paragraph{Preserving task evidence}
To ensure that transformed histories retain task-critical visual cues, we guide 
inversion with task-bearing spans from the language instruction. Let $\mathcal J$ 
denote all valid text token positions, and let $\mathcal G\subset\mathcal J$ denote 
the subset overlapping selected task-bearing spans, excluding special tokens. We 
extract frozen cross-attention weights from the video prior, normalized over 
$\mathcal J$ \emph{before} selecting $\mathcal G$:
\begin{equation}
 \begin{aligned}
 P^{bh}_{iu,j}&=\operatorname{softmax}_{j\in\mathcal J}
    (Q^{bh}_{iu}{K^{bh}_j}^{\!\top}/\sqrt d),\\
 a_{iu}&=\operatorname{mean}_{b,h}\sum_{j\in\mathcal G}P^{bh}_{iu,j},\\
 w_g(i,u)&=\frac{a_{iu}}{\sum_{i',u'}m_{i'u'}a_{i'u'}+\epsilon}.
 \end{aligned}
 \label{eq:task_attention_main}
\end{equation}
We average attention heads from blocks $\{6,12,18,24\}$ at noise level $\sigma=0.05$; 
the same blocks provide feature readouts $F_\psi$. We decode the transformed latent 
through the frozen decoder: $R(U)=D_{\rm Wan}(U)$. At valid locations $m_{iu}$, we 
minimize the task-weighted cosine distance $d(a,b)=1-a^\top b$:
\begin{equation}
 \mathcal L_{\rm pres}=
 \frac{\sum_{i,u}m_{iu}w_g(i,u)
 d(F_\psi(R(U))_{iu},F_\psi(\mathcal H_t)_{iu})}
 {\sum_{i,u}m_{iu}w_g(i,u)+\epsilon}.
 \label{eq:preserve}
\end{equation}
Source features and task weights are detached; gradients flow through the frozen 
decoder to update only the latent $U$.

\paragraph{Preserving dynamics}
We establish feature correspondences using mutual nearest-neighbor matching, inspired 
by video-editing consistency~\cite{tokenflow2023,fatezero2023}. A match $r=(iu,jv)$ 
is retained in $\mathcal C_{\rm src}$ if both endpoints are valid, their cosine 
similarity exceeds 0.8, and their cycle error is at most 1 feature-grid pixel. 
For each match, we define $D_r(H)=F_\psi(H)_{iu}-F_\psi(H)_{jv}$ and minimize:
\begin{equation}
 \mathcal L_{\rm dyn}=
 \frac{\sum_{r\in\mathcal C_{\rm src}}q_r
 \|D_r(R(U))-D_r(\mathcal H_t)\|_2^2}
 {\sum_{r\in\mathcal C_{\rm src}}q_r+\epsilon},
 \label{eq:dynamic}
\end{equation}
where $q_r\in[0,1]$ weights source-match confidence.

\paragraph{Guided partial inverse}
The frozen video prior uses the T2V path of Wan2.2-TI2V-5B~\cite{wan2025,wan22}, 
without a clamped first frame. From $Z_0=E_{\rm Wan}(\mathcal H_t)$, Euler 
integration moves toward partial noise before a bounded correction:
\begin{equation}
 \begin{aligned}
 U_{k+1}&=Z_k+(\sigma_{k+1}-\sigma_k)
      v_\psi(Z_k,\sigma_k;\ell),\\
 Z_{k+1}&=U_{k+1}-\eta_k\operatorname{Clip}_{b_k}
                         (\nabla_U\mathcal L_T(U_{k+1})),\\
 \mathcal L_T&=\mathcal L_{\rm pres}+\lambda_{\rm dyn}\mathcal L_{\rm dyn},
 \qquad \lambda_{\rm dyn}=1.
 \end{aligned}
 \label{eq:history_inversion}
\end{equation}
The eight-step inverse ends at $\tau_{\rm inv}=0.1$, with $\eta_k=0.01$ and 
$b_k=0.01\|U_{k+1}\|_2$ bounding each correction. The decoded history 
$\widetilde{\mathcal H}_t$ is re-encoded for the policy. A pair is retained only 
with finite values, at least 64 valid matches, $L_{\rm pres}\leq0.10$, and 
$L_{\rm dyn}\leq0.16$. Rejection disables transformed supervision, leaving the raw 
example. Fixed thresholds yield 82\% accepted coverage; Eq.~\ref{eq:student} defines 
its training weight.
\subsection{Unified Active Vision World--Action Modeling}
\label{sec:joint_model}

Having addressed the retention side of the retain--acquire problem through 
task-guided history inversion (Sec.~\ref{sec:inversion}), we now turn to the 
acquisition side. Arm actions alter the scene configuration, while head actions 
determine which spatial relations remain visible within the finite history window. 
We unify these mechanisms within a single world--action model that learns what 
evidence to retain through inversion-based training and what to acquire through 
executable head control.

\paragraph{View-aware history}
To enable the policy to distinguish camera motion from scene changes, we augment 
the history representation with camera-specific metadata. Building on 
EasyWAM-Unified~\cite{easywam2026}, we encode each camera's history independently 
through a frozen video VAE, then augment the visual tokens with camera identity, 
measured pose, relative exposure time, and validity flags. A temporal resampler 
aggregates these tokens into a fixed-size summary:
\begin{equation}
 \begin{aligned}
 V_t^c&=E_V(\mathcal H_t^c)+e_{\rm cam}^c+
                  e_{\rm time}^c+e_{\rm pose}^c+e_{\rm valid}^c,\\
 B_t&=G_\phi(\{V_t^c\}_c,\ell,s_t,\kappa,\delta_{\leq t}).
 \end{aligned}
 \label{eq:history_context}
\end{equation}
This summary $B_t$ provides recent evidence and its spatiotemporal context, 
allowing the policy to infer whether a cue has disappeared due to a head turn 
or due to manipulation. Current RGB retains a fine-grained clean-condition path 
for contact detection. At each decision, we recompute $B_t$ from the available observation window without carrying a recurrent visual state between decisions. History summaries pass through a DiT-width~\cite{dit2023} projector and join 
the context \emph{after} native text projection, retaining the pretrained 
language interface.

\paragraph{Observation actions inside the joint generator}
Rather than employing a separate gaze-selection module, we generate head actions 
$A^{\rm head}$ within the same world--action model that produces arm actions 
$A^{\rm arm}$. Head coordinates augment the native action vector before action 
projection. The model processes current clean tokens, noisy future-video tokens, 
and projected noisy action tokens with shared observed context. Separate output 
projections return video and action velocities:
\begin{equation}
 (\widehat v_y,\widehat v_{\rm arm},\widehat v_{\rm head})
       =f_\theta(Y_s,A_s,s;\,O_t,B_t,\ell,s_t),
 \label{eq:joint_velocity}
\end{equation}
where $O_t$ is the current clean visual condition and $s$ is flow time. Splitting 
arm and head outputs for loss normalization does not introduce separate experts 
or hierarchical planning. Both future and action tokens interact during joint 
denoising in training. This architecture provides future-video prediction as a 
co-training signal.

\begin{figure*}[!t]
 \centering
 \includegraphics[width=\textwidth]{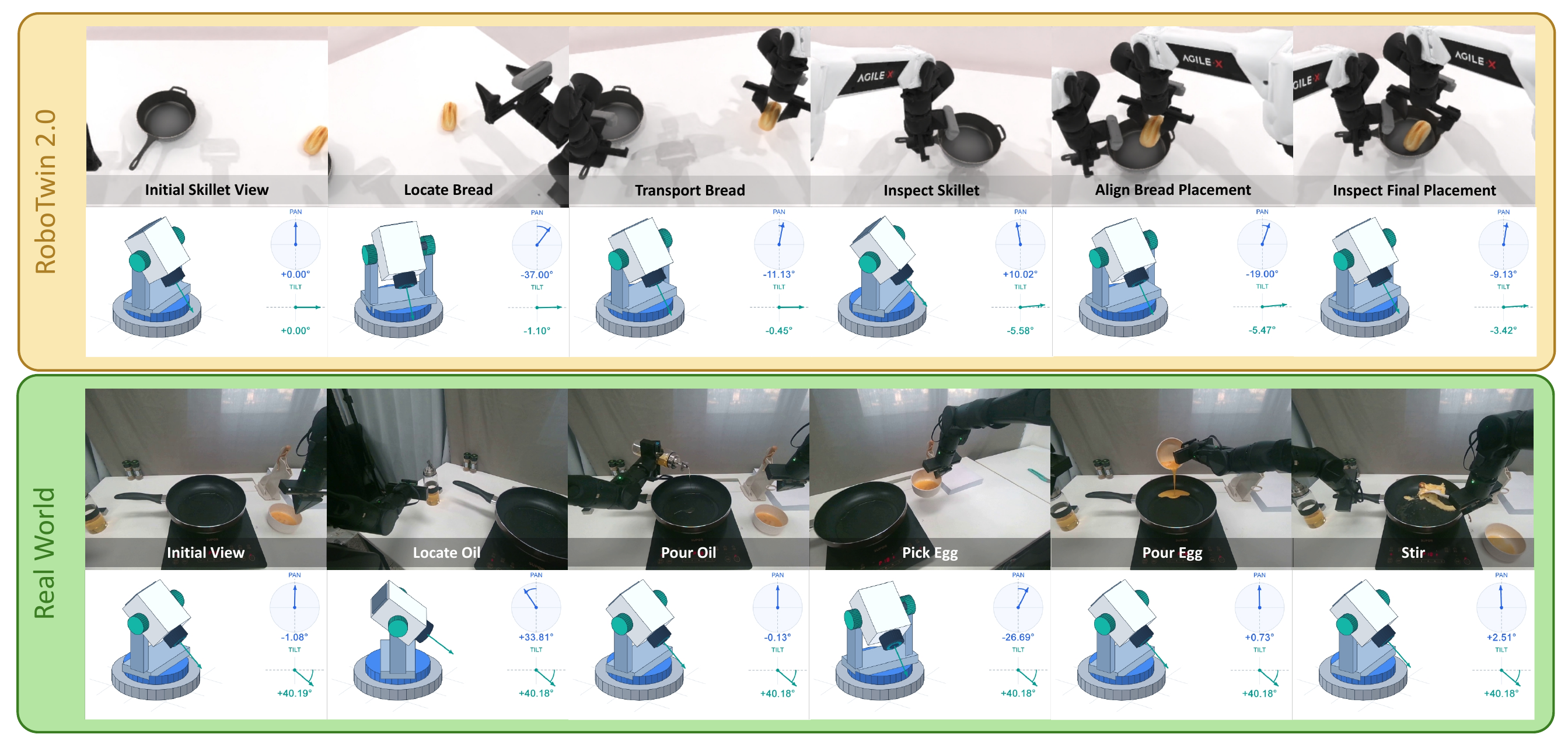}
 \caption{Representative head-view sequences with pan/tilt configurations: placing 
 bread in a pan in RoboTwin-AV (top); pouring oil and egg mixture, then stirring 
 on the real robot (bottom). ActiveWAM learns to expose task-critical spatial 
 relations (bread-to-skillet alignment, egg-to-pan trajectory) through head motion 
 while retaining previously acquired evidence within its history window.}
 \label{fig:tasks}
\end{figure*}

\paragraph{Active-vision demonstrations}
Demonstrations synchronize head and arm commands with RGB images recorded along
the executed trajectory. The RoboTwin-AV collector uses privileged task roles,
manipulation phase, visible geometry, instance masks, and depth. These signals
guide collection but are not provided to the policy. Evaluation follows native
task success predicates rather than agreement with the collector.
Future-video targets remain in their recorded camera streams to preserve the
visual effects of head motion. Demonstrations associate search with reacquisition,
approach with object--hand coverage, and placement with object--goal framing.
They also include stay, whose frequency alone does not establish a new observation
strategy. Inversion changes the observed condition while retaining the original
action and future-video targets, encouraging consistent decisions when
task-relevant cues survive.

\subsection{Paired Learning and Deployment}
\label{sec:execution}

Paired learning anchors raw and transformed histories to identical future-video and 
action targets, encouraging consistent decisions under nuisance view variations. Raw 
and transformed forward passes share all model weights and nonvisual metadata, 
differing only in their visual inputs. We encode the original current-plus-future 
clip once to obtain the shared target $Y_0$ and $A_0$. Following flow 
matching~\cite{flowmatching2022,rectifiedflow2022}, we interpolate with independent 
noise:
\begin{equation}
 Y_s=(1-s)Y_0+s\epsilon_y,\qquad
 A_s=(1-s)A_0+s\epsilon_a.
 \label{eq:flow}
\end{equation}
Paired passes share $Y_0$, $A_0$, flow time $s$, and noise samples $\epsilon_y,\epsilon_a$.

For each branch $b\in\{\mathrm{raw},\mathrm{stem}\}$, video, arm, and head errors 
are normalized and equally weighted:
\begin{equation}
 \mathcal L_{\rm FM}^b=\mathcal L_y^b+
          \mathcal L_{\rm arm}^b+\mathcal L_{\rm head}^b,
 \label{eq:fm}
\end{equation}
with velocity targets $\epsilon_y-Y_0$ and $\epsilon_a-A_0$. Valid pairs additionally 
align action velocities:
\begin{equation}
 \begin{aligned}
 \mathcal L_{\rm pair}
  &=\|\widehat v_a^{\rm raw}-\widehat v_a^{\rm stem}\|^2_{\rm valid},\\
 \mathcal L_{\rm WAM}
  &=\frac{\mathcal L_{\rm FM}^{\rm raw}
        +m_{\rm pair}\mathcal L_{\rm FM}^{\rm stem}}
         {1+m_{\rm pair}}
       +\lambda_{\rm pair}m_{\rm pair}\mathcal L_{\rm pair}.
 \end{aligned}
 \label{eq:student}
\end{equation}
We attempt pairing for 50\% of eligible windows and set $m_{\rm pair}=1$ only for 
accepted transformations, with $\lambda_{\rm pair}=0.05$.

\paragraph{Raw-history inference}
At deployment, the policy generates one action trajectory from raw history, executes 
a short prefix, and recomputes context from newly exposed frames. 
Deployment involves no online inversion, candidate ranking, or information-gain 
computation. A shared adapter enforces actuator limits and observation timing 
constraints. Historical frames retain their true ages, allowing the policy to reason 
about temporal coverage. Future-video latents predicted during training are auxiliary 
outputs; our action-only control confirms that 
deployment can omit future-video decoding without performance loss.

\section{Experiments}
\label{sec:experiment}

\paragraph*{Experimental questions}
Our experimental evaluation is organized around four central questions.
\textbf{(Q1)} Does task-guided inversion preserve task-relevant evidence across view changes?
\textbf{(Q2)} Do learned observation actions improve over fixed or rule-based control?
\textbf{(Q3)} What roles do video co-training and paired learning play?
\textbf{(Q4)} Can the system compose multi-stage physical tasks with acceptable efficiency?

\begin{table*}[!t]
 \centering
 \caption{Success on TAVIS (\%). The first row reproduces the official reported $\pi_0$ results. Our evaluations retain native demonstrations and sensor access and evaluate a fixed checkpoint under three environment seeds, with 96 episodes/task/seed per condition. ID/S/P denote in-distribution, spatial OOD, and initial-pose OOD. }
 \label{tab:main_benchmarks}
 \footnotesize
 \setlength{\tabcolsep}{0.6pt}
 \begin{tabular*}{\textwidth}{@{\extracolsep{\fill}}lrrrrrrrrrrrr@{}}
 \toprule
 &\multicolumn{3}{c}{Head / GR1T2}&\multicolumn{3}{c}{Head / Reachy2}
 &\multicolumn{3}{c}{Hands / GR1T2}&\multicolumn{3}{c}{Hands / Reachy2}\\
 \cmidrule(lr){2-4}\cmidrule(lr){5-7}\cmidrule(lr){8-10}\cmidrule(lr){11-13}
 Method&ID&S&P&ID&S&P&ID&S&P&ID&S&P\\
 \midrule
$\pi_0$ ~\cite{pi02024} & 47.1 & 27.9 & 1.9 & 38.8 & 22.1 & 11.2 & 70.1 & 49.0 & 14.6 & 76.7 & 51.0 & 38.2\\
$\pi_{0.5}$~\cite{pi052025} & \awSD{51.0}{1.0} & \awSD{28.7}{0.6} & \awSD{7.7}{0.6} & \awSD{44.0}{0.0} & \awSD{27.0}{1.0} & \awSD{16.3}{0.6} & \awSD{73.3}{1.2} & \awSD{54.3}{1.5} & \awSD{18.7}{0.6} & \awSD{\textbf{80.0}}{1.0} & \awSD{55.0}{1.0} & \awSD{43.7}{1.2}\\
Fast-WAM & \awSD{47.0}{2.0} & \awSD{30.0}{1.0} & \awSD{6.0}{1.0} & \awSD{41.0}{1.0} & \awSD{27.0}{2.0} & \awSD{17.0}{2.0} & \awSD{70.7}{1.2} & \awSD{51.0}{1.0} & \awSD{24.3}{1.5} & \awSD{76.0}{1.0} & \awSD{52.0}{2.6} & \awSD{41.3}{0.6}\\
EasyWAM-Unified & \awSD{51.3}{1.5} & \awSD{30.0}{2.0} & \awSD{11.7}{1.5} & \awSD{46.0}{1.0} & \awSD{27.3}{1.2} & \awSD{20.7}{1.2} & \awSD{68.0}{2.6} & \awSD{53.3}{0.6} & \awSD{23.7}{0.6} & \awSD{76.3}{0.6} & \awSD{52.7}{0.6} & \awSD{39.7}{2.9}\\
\textbf{ActiveWAM (ours)} & \awSD{\textbf{63.7}}{2.1} & \awSD{\textbf{47.0}}{1.0} & \awSD{\textbf{24.7}}{0.6} & \awSD{\textbf{52.7}}{1.5} & \awSD{\textbf{44.0}}{2.6} & \awSD{\textbf{31.3}}{1.5} & \awSD{\textbf{75.7}}{1.2} & \awSD{\textbf{64.0}}{0.0} & \awSD{\textbf{36.3}}{0.6} & \awSD{{79.7}}{3.1} & \awSD{\textbf{63.3}}{0.6} & \awSD{\textbf{54.7}}{1.5}\\
 \bottomrule
 \end{tabular*}
\end{table*}

\begin{table}[!t]
 \centering
 \caption{RoboTwin-AV SR (\%): 50 tasks, three seeds,
 100 episodes/task/seed/condition; shared sensor/action interfaces.
 App./Pose/Both: appearance/head-pose/compound shifts. }
 \label{tab:main_robotwin_av}
 \footnotesize
 \setlength{\tabcolsep}{1.8pt}
 \begin{tabular*}{\columnwidth}{@{\extracolsep{\fill}}lrrrr@{}}
 \toprule
 Method & Clean & App. & Pose & Both\\
 \midrule
ACT & \awSD{31.3}{1.5} & \awSD{8.7}{1.2} & \awSD{20.0}{1.0} & \awSD{3.7}{0.6}\\
Diffusion Policy & \awSD{34.7}{0.6} & \awSD{11.3}{1.2} & \awSD{16.3}{0.6} & \awSD{5.0}{1.0}\\
$\pi_{0.5}$ & \awSD{63.0}{1.0} & \awSD{42.3}{1.2} & \awSD{43.0}{1.7} & \awSD{34.0}{1.0}\\
Fast-WAM & \awSD{76.3}{1.2} & \awSD{41.7}{1.5} & \awSD{56.0}{1.0} & \awSD{33.3}{1.2}\\
EasyWAM-Unified & \awSD{75.0}{1.0} & \awSD{46.3}{1.5} & \awSD{56.0}{0.0} & \awSD{33.3}{1.2}\\
AV-ALOHA & \awSD{39.3}{1.2} & \awSD{25.7}{0.6} & \awSD{26.3}{0.6} & \awSD{19.3}{1.5}\\
\midrule
\textbf{ActiveWAM} & \awSD{\textbf{80.3}}{2.1} & \awSD{\textbf{60.7}}{0.6} & \awSD{\textbf{65.0}}{3.0} & \awSD{\textbf{53.3}}{1.5}\\
 \bottomrule
 \end{tabular*}
\end{table}

\begin{table}[!t]
 \centering
 \caption{RoboTwin-AV ablations (SR, \%). }
 \label{tab:main_mechanism}
 \footnotesize
 \setlength{\tabcolsep}{1.8pt}
 \begin{tabular}{p{0.55\columnwidth}cc}
 \toprule
 Variant & Clean & Compound\\
 \midrule
ActiveWAM (full) & \awSD{80.3}{2.1} & \awSD{53.3}{1.5}\\
w/o history & \awSD{73.7}{0.6} & \awSD{34.0}{0.0}\\
w/o inversion (raw history) & \awSD{76.7}{1.5} & \awSD{41.7}{1.5}\\
Same-Wan framewise inversion & \awSD{79.7}{0.6} & \awSD{43.5}{1.0}\\
w/o task preservation & \awSD{78.7}{0.6} & \awSD{41.0}{1.0}\\
w/o dynamic preservation & \awSD{82.0}{1.0} & \awSD{44.7}{1.2}\\
w/o pair consistency & \awSD{81.0}{1.0} & \awSD{44.7}{1.5}\\
w/o view metadata & \awSD{79.3}{2.3} & \awSD{46.3}{0.6}\\
Raw/raw equal-forward control & \awSD{76.7}{0.6} & \awSD{40.0}{1.0}\\
Task-safe augmentation & \awSD{76.0}{2.6} & \awSD{45.3}{0.6}\\
Task-weighted direct noising & \awSD{77.0}{1.0} & \awSD{46.0}{0.0}\\
VAE round trip & \awSD{76.0}{1.7} & \awSD{42.7}{0.6}\\
Source-feature auxiliary & \awSD{79.3}{0.6} & \awSD{45.0}{1.0}\\
 \bottomrule
 \end{tabular}
\end{table}

\begin{table}[!t]
 \centering
 \caption{Head control with inversion fixed. Clean and Pose give SR (\%);
 Co-vis.: object--goal co-visibility (\%); travel: rad/episode.}
 \label{tab:main_active_view}
 \footnotesize
 \setlength{\tabcolsep}{1.8pt}
 \begin{tabular}{p{0.31\columnwidth}cccc}
 \toprule
 Head controller & Clean & Pose & Co-vis. & Travel\\
 \midrule
Locked & \awSD{76.3}{0.6} & \awSD{45.0}{1.0} & \awSD{61.0}{1.0} & 0.0\\
Object tracking & \awSD{76.3}{0.6} & \awSD{57.3}{1.2} & \awSD{73.7}{0.6} & \awSD{4.1}{0.1}\\
Relation look-at + stay & \awSD{79.3}{0.6} & \awSD{61.0}{0.0} & \awSD{81.0}{1.0} & \awSD{2.5}{0.1}\\
Joint learned (ours) & \awSD{80.3}{2.1} & \awSD{65.0}{3.0} & \awSD{87.7}{0.6} & \awSD{1.7}{0.1}\\
+ matched grounding & \awSD{81.3}{0.6} & \awSD{62.7}{0.6} & \awSD{86.3}{3.8} & \awSD{1.9}{0.1}\\
 \bottomrule
 \end{tabular}
\end{table}

\subsection{Benchmarks and Experimental Setup}
\label{sec:data}

TAVIS~\cite{tavis2026} provides native Head/Hands trajectories with 96 test 
episodes per task, seed, and split. Its pose-OOD condition perturbs end-effector 
positions ($\sigma=10$\,cm) and neck yaw/pitch ($\sigma=10^\circ$) to test 
spatial robustness.

We extend RoboTwin~2.0~\cite{robotwin22025} with synchronized RGB observations 
and executable pan/tilt control across all 50 tasks, creating RoboTwin-AV. To 
generate active-vision demonstrations, we implement a gaze collector that uses 
instance masks and rendered depth to compute waypoints, with a 0.06-rad stay 
deadband to avoid unnecessary head motion. Collector-privileged inputs (masks, 
depth) are withheld at test time; success evaluation follows native task predicates. 
We train on 100 successful trajectories per task and evaluate on 100 test episodes 
per task, seed, and condition (clean, appearance, initial-head-pose, compound). Train/test splits are created before temporal windowing, 
and failures or timeouts are retained in the test set. Figure~\ref{fig:tasks} 
illustrates representative simulated and physical task sequences. For simulation results, each configuration is tested under three different seeds.

\paragraph{Baselines and additional benchmarks}
We compare against per-task baselines (ACT~\cite{act2023}, Diffusion 
Policy~\cite{dp2023}) and per-suite methods ($\pi_0$~\cite{pi02024}, Fast-WAM, 
EasyWAM-Unified). For fair comparison, all re-implemented baselines use identical 
RGB inputs, head--arm action interfaces, and native success predicates. Published 
benchmark scores (LIBERO~\cite{libero2023}, LIBERO-Plus~\cite{liberoplus2025}) 
are reported separately with their native interfaces.

\paragraph{Training and metrics}
We fine-tune DiT modules and the 16-token adapter where VAE, text encoder, and 
video prior are frozen. The policy receives nine RGB frames at $240\times320$ resolution. For 
ablation studies, control experiments match the full training recipe except the 
ablated component, with head trajectories held consistent across matched conditions. 

\subsection{Main Benchmark Comparisons}
\label{sec:standard_results}
\label{sec:active_results}

\paragraph{Results on TAVIS}
On TAVIS, ActiveWAM improves over the strongest listed baseline by up to 17.0 
percentage points (Table~\ref{tab:main_benchmarks}). The gains are pronounced 
under pose-OOD conditions: +13.0/+10.6\,pp on Head tasks and +12.0/+11.0\,pp on Hands 
tasks for GR1T2/Reachy2, respectively. These results show higher task success under end-effector and camera pose perturbations. We examine the contributions of history inversion and observation control through the controlled comparisons below. To characterize gaze timing, we measure gaze--action lead time on Head ID tasks. Mean 
lead times are 0.75\,s (GR1T2) and 0.78\,s (Reachy2), with 97\% of gaze actions 
classified as valid (878 and 763 successful episodes, respectively).

\paragraph{Results on RoboTwin-AV}
On RoboTwin-AV (Table~\ref{tab:main_robotwin_av}), ActiveWAM achieves gains of 
+4.0, +14.4, +9.0, and +19.3 percentage points over the strongest listed baseline 
under clean, appearance, pose, and compound shifts, respectively. Under compound shifts, ActiveWAM exceeds Fast-WAM by 20.0 percentage points. It also achieves 53.3$\pm$1.5\% success compared with 41.7$\pm$1.5\% for the raw-history baseline. The matched ablations below examine which components contribute to this improvement.

\paragraph{Fixed-camera generalization}
\suppressfloats[t]
\begin{table}[t]
 \centering
 \caption{Fixed-camera checks (SR, \%). LIBERO-Plus weights all
 10,030 variants by category counts.}
 \label{tab:main_generalization}
 \footnotesize
 \setlength{\tabcolsep}{1.8pt}
 \begin{tabular}{lrr}
 \toprule
 Variant & LIBERO & LIBERO-Plus\\
 \midrule
Raw history & \awSD{98.4}{0.8} & \awSD{68.2}{0.1}\\
Task-safe augmentation & \awSD{98.5}{0.2} & \awSD{72.5}{0.1}\\
Source-feature auxiliary & \awSD{98.7}{0.1} & \awSD{71.9}{0.2}\\
ActiveWAM & \awSD{98.7}{0.1} & \awSD{76.9}{0.1}\\
 \bottomrule
 \end{tabular}
\end{table}

We also evaluate fixed-camera benchmarks to examine whether the training approach remains useful when the policy cannot move the camera. On LIBERO-Plus, ActiveWAM achieves 76.9\% success compared with 68.2\% for raw history, 72.5\% for task-safe augmentation, and 71.9\% for source-feature auxiliary supervision (Table~\ref{tab:main_generalization}). On LIBERO, success is 98.7\% compared with 98.4\% for raw history. These results show an improvement on LIBERO-Plus while the reported LIBERO scores remain close. The interaction between inversion and head control is examined separately on RoboTwin-AV.
\subsection{Mechanism Validation and Control Experiments}
\label{sec:history_results}
\label{sec:interaction}

\begin{table}[!t]
 \centering
 \caption{RoboTwin-AV attribution/sensitivity (SR, \%): trajectory-consistent
 factorial data and matched training/inference graphs. Coverage:
 accepted/attempted inversion windows.}
 \label{tab:main_attribution}
 \footnotesize
 \setlength{\tabcolsep}{1.8pt}
 \begin{tabular}{llrr}
 \toprule
 Inversion & Head & Clean & Compound\\
 \midrule
 Off & Locked & \awSD{74.3}{0.6} & \awSD{23.7}{1.2}\\
 On & Locked & \awSD{76.3}{0.6} & \awSD{30.3}{0.6}\\
 Off & Active & \awSD{76.7}{1.5} & \awSD{41.7}{1.5}\\
 On & Active & \awSD{80.3}{2.1} & \awSD{53.3}{1.5}\\
 \midrule
 Video loss & Inference & Clean & Compound\\
 \midrule
 On & Joint & \awSD{80.3}{2.1} & \awSD{53.3}{1.5}\\
 Off & Joint & \awSD{75.0}{1.0} & \awSD{44.0}{0.0}\\
 On & Action-only graph & \awSD{81.0}{0.0} & \awSD{52.7}{0.6}\\
 Off & Action-only graph & \awSD{76.7}{1.2} & \awSD{44.7}{0.6}\\
 \midrule
 $\lambda_{\rm dyn}$ & Coverage (\%) & Clean & Compound\\
 \midrule
 0.0 & 86.0 & \awSD{82.0}{1.0} & \awSD{44.7}{1.2}\\
 0.1 & 85.0 & \awSD{82.0}{1.0} & \awSD{48.7}{1.2}\\
 0.5 & 84.0 & \awSD{81.3}{1.5} & \awSD{52.3}{1.2}\\
 1.0 & 82.0 & \awSD{80.3}{2.1} & \awSD{53.3}{1.5}\\
 2.0 & 80.0 & \awSD{80.0}{2.0} & \awSD{51.0}{2.0}\\
 \bottomrule
 \end{tabular}
\end{table}

\paragraph{Retention--acquisition interaction}
We cross history inversion with active head control to examine whether their benefits depend on one another (Table~\ref{tab:main_attribution}). Under compound shifts, success is 23.7\% with neither component, 30.3\% with inversion alone, 41.7\% with head control alone, and 53.3\% with both. The interaction estimate is approximately 5.0 percentage points, computed from the means displayed in the table. The corresponding estimates are approximately 1.6, $-0.3$, and 1.0 percentage points under clean, appearance, and pose conditions. This pattern suggests that inversion provides a greater benefit when head control is available under compound shifts.

\paragraph{History inversion and evidence preservation}
Removing task or dynamic preservation reduces compound success by 12.3 and 8.6 percentage points, respectively (Table~\ref{tab:main_mechanism}). ActiveWAM also outperforms framewise inversion using the same Wan prior, which achieves 43.5\% success. These results support the contribution of task guidance and temporal preservation to control under compound shifts.

\paragraph{Learned observation control and collector controls}
Table~\ref{tab:main_active_view}'s 20\,pp pose gain compares locked-camera to learned 
control, establishing the value of executable observation. Under an independently 
randomized collector, compound success remains 53\% (42\% raw); bounded external 
perturbations retain 51\%.
In 23\% of successful compound episodes, the policy maintains a stable view for at least 0.5\,s while manipulation continues. The learned controller records 1.70\,rad of head travel per episode and 65\% pose success, compared with 2.50\,rad and 61\% for relation look-at with stay, and 4.10\,rad and 57.3\% for object tracking (Table~\ref{tab:main_active_view}). With matched RGB grounding, the learned controller reaches 62.7\% pose success with 1.90\,rad of travel, compared with 61\% and 2.50\,rad for relation look-at with stay.

\begin{table}[!t]
 \centering
 \caption{Controller diagnostics. RoboTwin-AV pose-shift rows use matched RGB grounding. Co-vis./stale are percentages; time is timeout-capped reacquisition (10 s). Arms-busy is evaluated on the corresponding TAVIS slice; the oracle is an upper bound. }
 \label{tab:behavior}
 \footnotesize
 \setlength{\tabcolsep}{1.0pt}
 \begin{tabular}{@{}p{0.25\columnwidth}rrrrr@{}}
 \toprule
 Controller & \shortstack[r]{SR\\(\%)} & \shortstack[r]{Co-vis.\\(\%)} & \shortstack[r]{Time\\(s)} & \shortstack[r]{Travel\\(rad)} & \shortstack[r]{Stale\\(\%)}\\
 \midrule
Locked head & \awSD{45.0}{1.0} & \awSD{61.0}{1.0} & 10.0 & 0.0 & \awSD{1.2}{0.2}\\
Target centering & \awSD{57.3}{1.2} & \awSD{73.7}{0.6} & \awSD{2.3}{0.1} & \awSD{4.1}{0.1} & \awSD{8.2}{0.3}\\
Relation + stay & \awSD{61.0}{0.0} & \awSD{81.0}{1.0} & \awSD{1.9}{0.1} & \awSD{2.5}{0.1} & \awSD{5.1}{0.1}\\
Joint learned & \awSD{62.7}{0.6} & \awSD{86.3}{3.8} & \awSD{1.6}{0.1} & \awSD{1.9}{0.1} & \awSD{4.2}{0.3}\\
No extra sense arm & \awSD{62.0}{1.7} & \awSD{88.0}{1.0} & \awSD{1.6}{0.0} & \awSD{1.9}{0.1} & \awSD{4.2}{0.2}\\
Arms busy & \awSD{54.7}{0.6} & \awSD{76.3}{1.2} & \awSD{2.1}{0.1} & \awSD{2.4}{0.1} & \awSD{6.2}{0.2}\\
View oracle & \awSD{76.7}{0.6} & \awSD{91.7}{1.5} & \awSD{1.2}{0.1} & \awSD{1.8}{0.0} & \awSD{3.3}{0.3}\\
 \bottomrule
 \end{tabular}
\end{table}

\paragraph{Controller diagnostics}
Under matched RGB grounding, the learned controller achieves 62.7\% pose success compared with 57.3\% for target centering (Table~\ref{tab:behavior}). It also records higher co-visibility, lower head travel, and shorter timeout-capped reacquisition time, with values of 86.3\%, 1.90\,rad, and 1.6\,s compared with 73.7\%, 4.10\,rad, and 2.3\,s. The arms-busy result is evaluated on a separate TAVIS slice. The view oracle provides a reference result of 76.7\% success.

\paragraph{Prediction and paired-learning controls}
\label{sec:prediction_control}
Table~\ref{tab:main_attribution} crosses video loss and inference mode. Video 
co-training improves matched joint graphs from 44.0\% to 53.3\% compound SR; 
separately trained action-only graphs reach 52.7\% with video loss, 44.7\% without.
Thus video prediction is effective as a training signal but not required at test 
time. Decode-off diagnostics show zero action difference and 53\% SR over ten 
evaluations. Equal-forward and pair-consistency removal underperform 
(Table~\ref{tab:main_mechanism}).

\subsection{Composed Real-Robot Tasks}
\label{sec:real_results}
\label{sec:efficiency}

\begin{table}[!t]
 \centering
 \caption{Physical kitchen successes out of 20 attempts per task (60 per method and 300 across all five methods). Stage~1: first subtask only; Total: both subtasks completed in the same trial.}
 \label{tab:main_real}
 \footnotesize
 \setlength{\tabcolsep}{3pt}
 \begin{tabular}{lcccccc}
 \toprule
 & \multicolumn{2}{c}{Cucumber} & \multicolumn{2}{c}{Egg} & \multicolumn{2}{c}{Mixture}\\
 \cmidrule(lr){2-3}\cmidrule(lr){4-5}\cmidrule(lr){6-7}
 Method & S1 & Total & S1 & Total & S1 & Total\\
 \midrule
 $\pi_{0.5}$~\cite{pi052025} & 8 & 2 & 10 & 7 & 12 & 11\\
 Fast-WAM & 7 & 1 & 11 & 6 & 11 & 7\\
 EasyWAM-Unified & 7 & 1 & 10 & 7 & 12 & 9\\
 Ours w/o inversion & 10 & 4 & 12 & 7 & 14 & 11\\
 ActiveWAM (full) & 11 & 7 & 15 & 11 & 14 & 12\\
 \bottomrule
 \end{tabular}
\end{table}

We deploy ActiveWAM on an AirbotPlay fixed-base dual-arm platform with a 
fixed focal Realsense D455 RGB camera on an IQR pan/tilt 2-DOF head. Each of three task 
segments composes two separately trained policies: (i) \emph{slicing}—place 
cucumber horizontally, retrieve knife, then slice; (ii) \emph{egg frying}—pour 
oil, crack egg while the left arm grasps the pan handle, then retrieve spatula 
and stir; (iii) \emph{mixing}—pour cucumber and cooked egg, then retrieve spatula 
and stir-fry. Egg stirring retains the pan-handle grasp; stage switching is 
prescribed.
Human inter-segment actions enter neither evaluation nor montage. 
Table~\ref{tab:main_real} reports Stage~1 and total-task success
for the same 20 complete trials per task and method, yielding
60 trials per method and 300 trials across the five methods.
ActiveWAM achieves 66.7\%/50.0\% Stage~1/total-task success,
versus 60.0\%/36.7\% without inversion and 48.3\%/23.3\%
for Fast-WAM.
\subsection{Deployment Efficiency and Failure Analysis}
\label{sec:deployment}

On an NVIDIA RTX 4090 GPU, action inference takes approximately 165\,ms. Action chunking yields an effective rate near 15\,Hz, while interpolation maintains 25\,Hz commands.

ActiveWAM completes the full task in 30 of 60 physical trials.
We review all 30 failed trials and assign each to one primary
category: manipulation errors (12 trials, 40.0\%), premature
termination (10 trials, 33.3\%), targets outside the pan/tilt
camera's reachable field of view (5 trials, 16.7\%), and
hardware issues (3 trials, 10.0\%). These percentages are
calculated over the 30 failed trials. The categories describe
the failure modes without establishing
underlying causes.

\section{Conclusion}
\label{sec:conclusion}

ActiveWAM formulates active vision manipulation as evidence-aware retain--acquire 
control, where a policy must preserve task-critical cues within a finite history 
window while controlling which evidence it acquires next. Task-guided history 
inversion preserves task-relevant evidence and visible temporal changes during 
training, while unified world--action generation produces executable bimanual and 
head actions from raw observations at deployment. Across TAVIS, RoboTwin-AV, and 
physical tasks, the method improves over strong baselines by up to 19.3 percentage points under compound distribution shifts. Matched factorial and collector controls 
confirm that evidence retention and active observation are complementary. The current system uses a fixed-focal-length head, short 
history window, and prescribed stage switching; natural extensions include wider 
fields of view, optical zoom, end-to-end segmentation, and longer-horizon memory 
mechanisms.

\label{layout:references}
\bibliographystyle{IEEEtran}

\balance
\bibliography{references}

\end{document}